\documentclass[conference]{IEEEtran}
\IEEEoverridecommandlockouts
\usepackage{cite}
\usepackage{amsmath,amssymb,amsfonts}
\usepackage{algorithmic}
\usepackage{graphicx}
\usepackage{caption}
\usepackage{textcomp}
\usepackage{xcolor}
\usepackage{booktabs}
\usepackage{multirow}
\usepackage{adjustbox}
\def\BibTeX{{\rm B\kern-.05em{\sc i\kern-.025em b}\kern-.08em
    T\kern-.1667em\lower.7ex\hbox{E}\kern-.125emX}}
\begin{document}

\title{\vspace{0.25in}StenoVLA-3D: 3D-Aware Reasoning VLA for Navigation Through Gastrointestinal Stenoses\\

}
\author{
Tamima Tabassum\textsuperscript{*}, Yiming Huang\textsuperscript{*},
Tianchun Wu, Changjing Liu, Zhiqing Tang,\\
Chikit Ng, Beilei Cui, Liangjing Shao, Jiewen Lai,
and Hongliang Ren\textsuperscript{\(\dagger\)}\\[2pt]
\small{\textsuperscript{*}Equal contributor, 
\textsuperscript{\(\dagger\)}Corresponding author
}
}
\maketitle

\begin{abstract}
Autonomous endoscopic navigation requires the policy model to predict actions from texture-poor monocular       
  observations, make safe control decisions, and retain evidence of lesions after they leave the field of view.
  Existing vision-language-action (VLA) models primarily rely on visual appearance and short-term context,     
  limiting geometric grounding and episode-level reporting. We introduce StenoVLA-3D, a 3D-aware VLA     
  framework for navigating through stenotic regions. We integrate point-maps into the Cosmos-Reason 2 backbone through learned geometry-gated fusion, and also propose a temporal state branch to model traversal progress. Our reasoning-and-action backbone predicts grounded reasoning with actions, while dedicated heads estimate stenosis shape and generate the final lesion report. We further introduce EndoCausal, an episode-level dataset with lesion annotations, actions, and temporally grounded reasoning. On 40 held-out recorded test episodes, StenoVLA-3D reaches 95.2\% semantic accuracy and 83.4\% action accuracy. On the physical 3-DoF endoscope, it attains 88.9\% and 77.8\% task success in esophageal and colonic phantoms (36 trials each), substantially outperforming the evaluated baselines.  
\end{abstract}

\begin{IEEEkeywords}
Robotic Endoscopy, Vision-Language-Action Models, Autonomous Navigation.
\end{IEEEkeywords}

\section{Introduction}
Gastrointestinal stenosis and strictures pose challenges for endoscopic assessment and intervention because luminal narrowing restricts access and complicates instrument positioning. The ASGE report \cite{akshintala2025tools} emphasizes evaluating stricture location, length, diameter, and etiology before treatment, highlighting the need to understand both geometry and clinical context. Safe instrument–tissue interaction is another important concern. \cite{del2025soft} developed a soft robot that monitors contact forces and redistributes excessive pressure during colonoscopy, illustrating the importance of tissue-aware robotic assistance.

Detecting a narrowed opening alone is insufficient, as surrounding tissue and lesions also affect clinical decisions. Robotic navigation therefore requires interpreting changing views, maintaining lumen alignment, tracking traversal stages, and retaining lesion information for reporting.



Existing VLAs remain limited for clinical endoscopic navigation. EndoVLA \cite{ng2025endovla} focuses on visual tracking without explicit reasoning, while general-purpose VLAs~\cite{intelligence2025pi05visionlanguageactionmodelopenworld} primarily generate actions from observations. Clinical applications, however, require interpretable reasoning and 3D anatomical information to understand complex spatial relationships.

\begin{figure*}[t]
    \centering
    \includegraphics[width=1.0\textwidth]{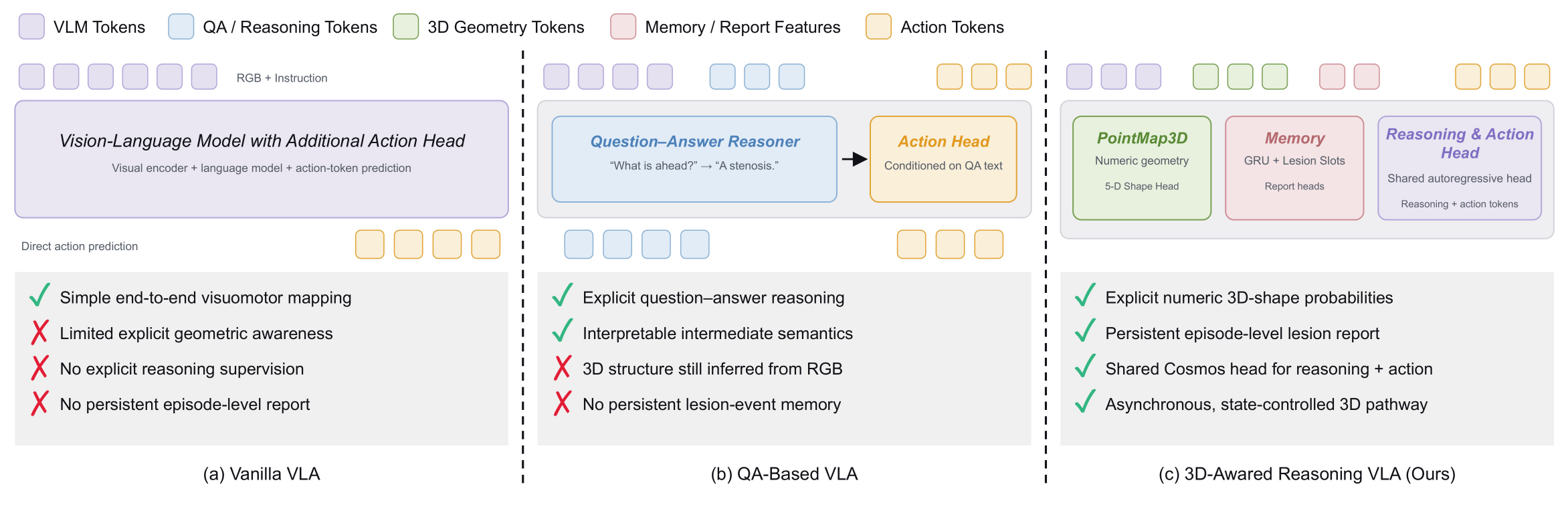}
    \caption{Comparison of different VLA paradigms. Our 3D-aware reasoning VLA integrates explicit 3D geometry and persistent lesion memory, facilitating the VLA for stronger reasoning and action prediction.}
    \label{compare}
\end{figure*}

We introduce StenoVLA-3D, a 3D-aware VLA framework that studies these capabilities in a controlled phantom setting. Our model generates structured reasoning, navigation states, opening-shape predictions, and discrete motion actions. A deterministic controller activates or deactivates DA3 reconstruction according to the predicted navigation state, and available point-map features condition subsequent predictions. Temporal and persistent lesion memories accumulate evidence for an episode-level report of lesion types, stenosis presence, and opening shape.
Fig.~\ref{compare} contrasts this organization with two common VLA designs. Vanilla VLAs map visual tokens directly to actions. QA-based VLAs emit intermediate language, but they do not keep persistent lesion-event memory or an explicit 3D pathway. StenoVLA-3D adds numeric point-map geometry, dual-timescale memory for reporting, a shared reasoning-and-action head, and asynchronous reconstruction gated by the predicted navigation state.
Our contributions are:
\begin{itemize}
    \item A 3D-aware reasoning VLA framework combining point-map encoding, temporal and persistent memory for navigation action prediction, reasoning, shape estimation and episode-level reporting.
    \item EndoCausal, a dataset of 400 phantom sequences connecting endoscopic observations with lesion annotations, navigation actions, reconstruction-control targets, and temporally grounded reasoning.
    \item A controlled evaluation of geometric representations and memory components, with StenoVLA-3D achieving task success rates of 88.9\% in esophagoscopy and 77.8\% in colonoscopy,
    complemented by evaluations under appearance and deformation shifts. 
\end{itemize}

\section{Related Work}

\subsection{Vision-Language-Action Models}
VLA models adapt pretrained multimodal representations for robot control, with performance influenced by backbone selection and policy architecture, as examined in RoboVLMs \cite{li2026matters}. Reasoning-augmented approaches introduce intermediate representations before action prediction: ECoT \cite{michal2024robotic} and EMMA-X \cite{sun2025emma} ground textual reasoning in observations and spatial relationships, whereas CoT-VLA \cite{zhao2025cot} generates future images as visual subgoals. Complementary work addresses information unavailable in the current observation. MemoryVLA \cite{shi2026memoryvla} retrieves perceptual details and semantic information from a memory bank, while OptimusVLA \cite{li2026global} combines task-level trajectory priors with recent-action memory to improve temporal consistency. In visuomotor control, HALO \cite{shah2026memory} learns task-relevant retrieval from observation histories for long-horizon tasks under partial observability. Building on these approaches, we use recent visual context to support endoscopic navigation and persistent memory to retain lesion evidence for episode-level reporting after lesions leave the field of view.

\subsection{Autonomous Endoscopic Navigation}
Autonomous endoscopic navigation has been explored through visual servoing, supervised learning, and reinforcement learning (RL). Lu et al. \cite{lu2023autonomous} combine monocular depth guidance with endoscope-body shape planning, while Hwang et al. \cite{hwang2026autonomous} learn steering targets and collision detection for robotic colonoscopy through benchtop and porcine evaluation. BronchoCopilot \cite{zhao2024bronchocopilot} combines visual and pose information for RL based navigation in simulated airways. For wireless capsules, edge–contour–depth fusion \cite{wu2026transferable} supports anatomical landmark-based gastric navigation and simulation-to-ex-vivo transfer. More recently, EndoVLA \cite{ng2025endovla} introduces language-conditioned endoscopic target tracking through supervised and reinforcement fine-tuning. However, it is limited for navigating to the narrowed stenoses without the 3D knowledge. Our work studies this complementary setting through geometry-informed action prediction during stenosis traversal and persistent lesion memory for episode-level reporting, evaluated on recorded gastrointestinal phantom episodes.



\begin{figure*}[t]
    \centering
    \includegraphics[width=1.0\textwidth]{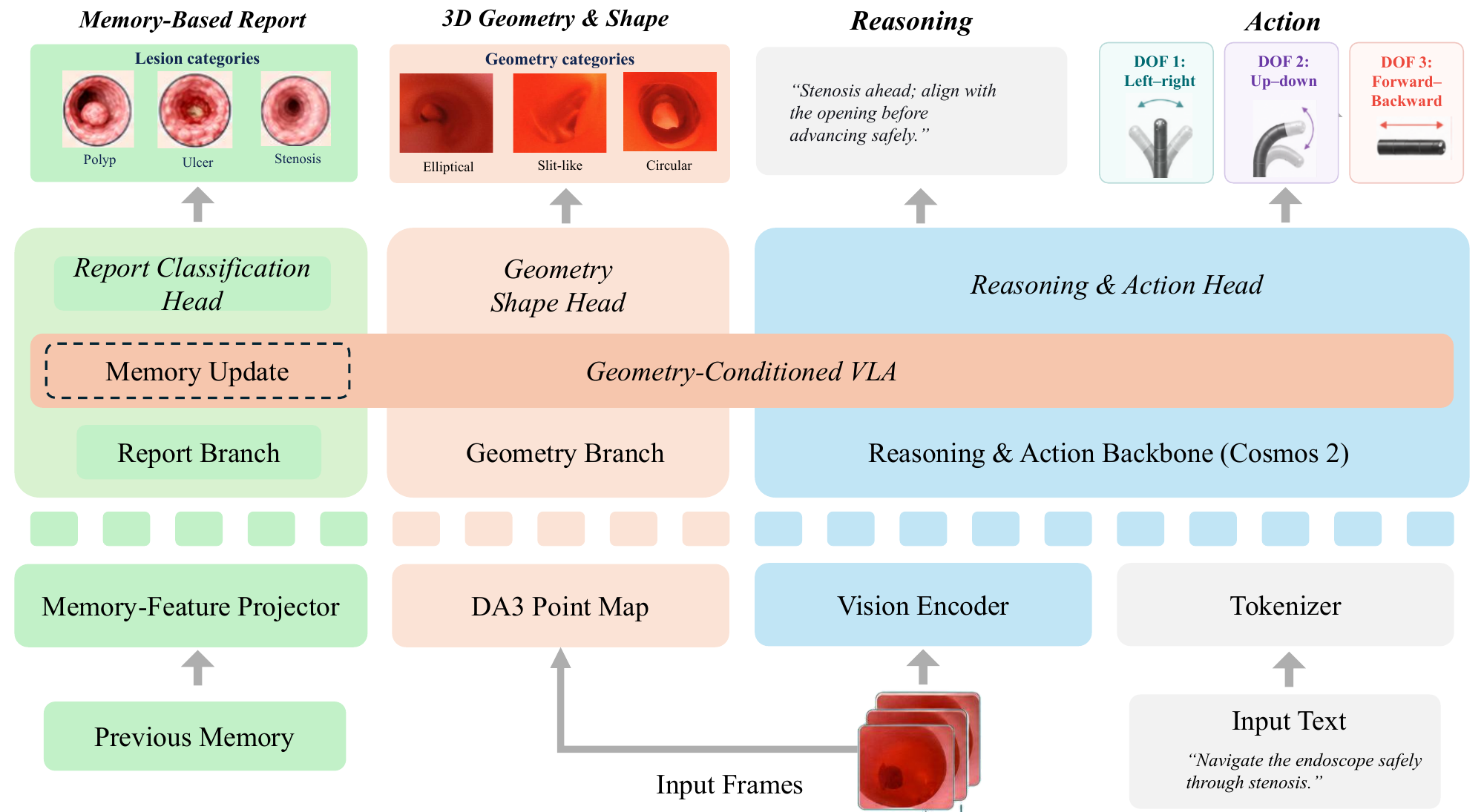}
    \caption{Overview of the proposed 3D-aware reasoning VLA. The framework fuses persistent state memory, DA3-derived 3D geometry, visual observations, and language instructions through geometry-conditioned VLA. Specialized heads predict reports and geometric shapes, while the shared Cosmos-Reason2~\cite{nvidia2025cosmosreason1physicalcommonsense} backbone generates semantic reasoning and 3-DoF actions.}
    \label{pipeline}
\end{figure*}

\section{Methodology}
\subsection{Preliminaries}
\label{sec:prelim}
Decisions are indexed by \(k=1,\ldots,K\), with \(K\) episode step.
On 20\,fps recordings the policy acts every fourth frame,
\(\rho_k=1+4(k-1)\).
The observation is
\(\mathcal{O}_k=\bigl(I_{\rho_k},\,\mathcal{H}_k,\,\mathcal{U},\,s^r_{k-1},\,v^g_k,\,\mathbf{z}_{g,k}\bigr)\),
where \(I_{\rho_k}\) is the current image,
\(\mathcal{H}_k\) is the visual history frames,
\(\mathcal{U}\) is the task instruction,
\(s^r_{k-1}\in\{\mathrm{ON},\mathrm{OFF}\}\) is the reconstruction controller state,
\(v^g_k\in\{0,1\}\) is 1 if the point map is available,
and \(\mathbf{z}_{g,k}\in\mathbb{R}^{256}\) is the encoded geometry.
The VLA \(\pi_{\theta}\) maps \(\mathcal{O}_k\) to
\begin{equation}
\mathbf{Y}_k
=
\pi_{\theta}(\mathcal{O}_k)
=
\bigl(R_k,\, n_k,\, c_k,\, o_k,\, a_k,\, \hat{a}^r_k\bigr),
\end{equation}
which includes reasoning \(R_k\), navigation state \(n_k\), safety constraint \(c_k\),
opening-shape text \(o_k\), motion \(a_k\in\mathcal{A}\), and an auxiliary
reconstruction command
\(\hat{a}^r_k\in\{\textit{no\_change},\,\textit{activate\_3d},\,\textit{deactivate\_3d}\}\). The set \(\mathcal{A}\) comprises the eight in-plane translations, axial \(\textit{forward}/\textit{backward}\), and \(\textit{stop}\). \(\textit{backward}\) is controlled withdrawal under wall proximity or poor visibility. The objective is to identify lesions, align with and traverse a stenosis, terminate safely, and summarize the observed findings.

\subsection{Dataset Curation}
\label{sec:data}
We introduce the EndoCausal dataset which annotates the fields of \(\mathbf{Y}_k\) in
Sec.~\ref{sec:prelim}, together with lesion-presence and memory-write
labels and an episode-level report of lesion types and stenosis presence. Data were collected using a four-motor device controlling three endoscope degrees of freedom: yaw, pitch, and axial translation. One motor controls left--right steering, another controls up--down steering, and two counter-rotating motors jointly control forward and backward translation. The dataset comprises 400 sequences collected at 20 fps.
All frames and derived variants from the same original recording
are assigned to a single split to prevent information leakage. Instance annotations are reviewed convex hulls of stenoses, polyps,
and ulcers. Temporal events mark the usable-view boundary, stenosis confirmation, reconstruction start, stenosis entry, optional distal-end confirmation, reconstruction stop, and complete or incomplete termination. 

SAM~3~\cite{carion2026sam} was used to generate initial segmentation masks
for stenoses, polyps, and ulcers. The masks were manually reviewed and corrected before conversion into convex-hull annotations. Manual episode events identified the usable-view boundary, stenosis confirmation, reconstruction start, stenosis entry, optional distal-end confirmation, reconstruction stop, and complete or incomplete termination. Let \((x^s_k,y^s_k)\) be a reviewed stenosis centroid and
\((x_c,y_c)\) the image centre, with image \(x\) rightward and \(y\)
downward. Unless a recovery or termination event overrides the rule, the motion target is \(\textit{forward}\) if
\(|x^s_k-x_c|\le\tau_a\) and \(|y^s_k-y_c|\le\tau_a\)
(\(\tau_a=40\) pixels), and otherwise the 8-connected in-plane
action that reduces this displacement. Opening shape \(o_k\) is an episode-level qualitative class (\textit{circular, elliptical, slit-like, crescent, or irregular}), or \textit{not observable}.
It is not a calibrated metric of stenosis geometry.
These labels supply the structured opening-shape targets used in training and are assigned before geometry modality dropout.
Reasoning targets \(R_k\) follow an
observation--interpretation--decision template and must not use
information from future frames.

\subsection{3D-Aware Geometry Encoding}
\label{sec:geometry}
We utilize the frozen DA3-Base model~\cite{lin2025depth} to estimate relative multi-view geometry information. We utilize precomputed point maps for training, and at the evaluation stage we estimate the point maps asynchronously from available observations. Each view is stored as \(\mathbf{P}_i=[X_i,Y_i,Z_i,M_i]\), with validity mask \(M_i\) to mask the low-confidence points. XYZ values are scaled by the 95th percentile of valid positive depth in that view.
At step \(k\) we pack 16 uniformly sampled maps, repeating the
earliest map if fewer are available:
\begin{equation}
\mathbf{P}_k\in\mathbb{R}^{4\times 16\times 112\times 112},
\qquad
\mathbf{z}_{g,k}=f_{\psi}(\mathbf{P}_k)\in\mathbb{R}^{256}.
\end{equation}
Here \(\mathbf{P}_k\) is the packed point-map clip and \(f_{\psi}\) is a frozen 3D convolutional encoder inspired by~\cite{Tran_2015_ICCV}.

\subsection{Geometry-Conditioned VLA}
\label{sec:vla}

Fig.~\ref{pipeline} overviews the architecture. The VLA is initialized from Cosmos-Reason2-2B~\cite{nvidia2025cosmosreason1physicalcommonsense}
using the Qwen3-VL~\cite{bai2025qwen3} conditional-generation
implementation.
We freeze the pretrained backbone parameters, while updating the rank-16 LoRA adapters~\cite{hu2021lora} in the query,
key, value, and output projection layers.

Let \(\mathbf{H}_{k,t}\in\mathbb{R}^{d}\) denote a final-layer
backbone token, with \(d\) the hidden size of Cosmos-Reason2-2B.
The geometry embedding is projected to the same dimension:
\begin{equation}
\mathbf{q}_k=\phi(\mathbf{z}_{g,k})\in\mathbb{R}^{d}.
\end{equation}
The projection uses layer normalization, a linear layer, and GELU.
A feature-wise gate is computed from the first backbone token
and the projected geometry:
\begin{equation}
\mathbf{g}_k=
\sigma\left(
W_g\operatorname{LN}
([\mathbf{H}_{k,0};\mathbf{q}_k])+\mathbf{b}_g
\right).
\end{equation}
Geometry conditions the final-layer token representations with:
\begin{equation}
\widetilde{\mathbf{H}}_{k,t}
=
\mathbf{H}_{k,t}
+
v^g_k(\mathbf{g}_k\odot\mathbf{q}_k).
\end{equation}
The language output operates on these conditioned
representations to generate the structured response.
Geometry is therefore introduced as a final-layer residual rather than as additional input tokens to the backbone. We use gated late-residual conditioning to inject a compact
geometry embedding without altering the pretrained token structure. This provides global geometric context but not explicit token-level spatial correspondence. During training, geometry is randomly omitted with probability 0.5 for augmentation. The language-model target for \(o_k\) becomes \textit{not observable} without geometry.

\subsection{Temporal and Persistent Memory}
\label{sec:memory}

\noindent\textbf{Temporal memory.}
Let \(\mathcal{I}_k\) index image tokens in the prompt and let
\(t^{\mathrm{end}}_k\) identify the final prompt token.
The memory feature is defined as:
\begin{equation}
\mathbf{e}_k =
F_{\mathrm{proj}}\left(
\left[
\frac{1}{|\mathcal{I}_k|}
\sum_{t\in\mathcal{I}_k}\widetilde{\mathbf{H}}_{k,t};
\widetilde{\mathbf{H}}_{k,t^{\mathrm{end}}_k}
\right]
\right)
\in\mathbb{R}^{512},
\end{equation}
where \(F_{\mathrm{proj}}\) comprises layer normalization,
a linear projection, GELU, and dropout.
The temporal state is updated by GRU:
$\mathbf{h}_k=
\operatorname{GRU}(\mathbf{e}_k,\mathbf{h}_{k-1}),
\
\mathbf{h}_k\in\mathbb{R}^{512}.$ Let \(\mathcal{J}=\{\text{polyp},\text{ulcer},\text{stenosis}\}\).
Current lesion logits are obtained from the current feature
and updated temporal state by:
$\mathbf{d}_k=W_d[\mathbf{e}_k;\mathbf{h}_k]+\mathbf{b}_d
\in\mathbb{R}^{3},$
with one logit \(d_{k,j}\) per class \(j\in\mathcal{J}\), and the lesion head weight $W_d$. The temporal state supports persistent-memory updates and report prediction.\\

\noindent\textbf{Persistent lesion memory.}
The memory state is defined as:
\begin{equation}
\mathcal{M}_k
=
\bigl(
\mathbf{h}_k,
\{\mathbf{s}_{k,j}\}_{j\in\mathcal{J}},
\mathbf{p}_k
\bigr),
\end{equation}
where \(\mathbf{s}_{k,j}\in\mathbb{R}^{512}\) is a lesion-class
slot and \(\mathbf{p}_k\in[0,1]^3\) contains accumulated
presence scores. All states are initialized to zero at
the beginning of an episode. A candidate slot \(\widetilde{\mathbf{s}}_{k,j}\) is a \(\tanh\)-linear map of
\([\mathbf{e}_k;\mathbf{h}_k]\).
The write logit \(w_{k,j}\) is a linear map of
\([\mathbf{e}_k;\mathbf{h}_k;\mathbf{s}_{k-1,j};p_{k-1,j}]\).
The update strength is
$\gamma_{k,j}=
\sigma(d_{k,j})\sigma(w_{k,j}),
\ j\in\mathcal{J}.$
The slot and presence score are updated as:
\begin{align}
\mathbf{s}_{k,j}
&=
(1-\gamma_{k,j})\mathbf{s}_{k-1,j}
+
\gamma_{k,j}\widetilde{\mathbf{s}}_{k,j},\\
p_{k,j}
&=
\max(p_{k-1,j},\gamma_{k,j}).
\end{align}
These states retain lesion evidence as persistent memory after the lesion leaves the current view.

\subsection{Shape Estimation, Reporting, and Reasoning}
\label{sec:outputs}
\label{sec:reasoning}

\noindent\textbf{3D shape estimation.}
When geometry is available, a linear head maps the point-map
embedding to five-class logits
\begin{equation}
\boldsymbol{\ell}_k=W_o\mathbf{z}_{g,k}+\mathbf{b}_o\in\mathbb{R}^{5}
\end{equation}
over \(\mathcal{C}_o=\{\textit{circular}, \textit{elliptical},
\textit{slit-like}, \textit{crescent}, \textit{irregular}\}\).
\(\mathcal{L}_{\mathrm{shape}}\) supervises this head only on
geometry-active steps. Each VLA output \(\mathbf{Y}_k\) also contains a text field \(o_k\), whose training target is the episode-level class in \(\mathcal{C}_o\) when geometry is present and \textit{not observable} otherwise.
The final shape estimation copies the last such text that is a class in \(\mathcal{C}_o\):
\begin{equation}
k^\star=\max\{k\leq K:o_k\in\mathcal{C}_o\}.
\end{equation}
\(k^\star\) is the latest decision at which the VLA named a
valid opening shape rather than \textit{not observable}, and the
reported shape is \(o_{k^\star}\). If no decision satisfies the
constraint, the output writes \textit{not observable}.\\

\noindent\textbf{Episode-level report.}
At episode completion, the report network takes the concatenated input:
\begin{equation}
\mathbf{r}_K=
[
\mathbf{h}_K;
\operatorname{vec}(\mathbf{S}_K);
\mathbf{p}_K
]
\in\mathbb{R}^{2051},
\end{equation}
where \(\mathbf{h}_K\) is the final GRU state,
\(\mathbf{S}_K=[\mathbf{s}_{K,j}]_{j\in\mathcal{J}}\)
stacks the three persistent slots,
\(\operatorname{vec}(\cdot)\) is the flatten operation,
and \(\mathbf{p}_K\in[0,1]^3\) is the accumulated presence.
A shared network maps this vector to a 512-dimensional feature.
Separate classification heads predict polyp/ulcer presence and
stenosis presence. Sigmoid probabilities are thresholded at 0.5
and inserted into a fixed report template together with
\(o_{k^\star}\).\\

\noindent\textbf{Step-level reasoning.}
At each decision, the language-model head generates the structured
fields of \(\mathbf{Y}_k\) in Sec.~\ref{sec:prelim} in order: a
temporally grounded reasoning trace \(R_k\), then the navigation
state, safety constraint, opening shape, motion action, and
reconstruction command. The trace follows an
observation-interpretation-decision structure and is supervised by \(\mathcal{L}_{\mathrm{LM}}\). It uses evidence available up to time \(k\). 

\subsection{Training}
\label{sec:training}

Training proceeds in two stages. First, the pretrained PointMap3D
encoder and VLA backbone are frozen. We optimize the LoRA
adapters, geometry projection, gating modules, memory-feature
projector, temporal GRU, persistent lesion memory, lesion heads,
and report network. The objective is:
\begin{equation}
\begin{aligned}
\mathcal{L}
&=
\lambda_{\mathrm{LM}}\mathcal{L}_{\mathrm{LM}}
+
\lambda_{\mathrm{lesion}}\mathcal{L}_{\mathrm{lesion}}
+
\lambda_{\mathrm{write}}\mathcal{L}_{\mathrm{write}} \\
&\quad
+
\lambda_{\mathrm{report}}\mathcal{L}_{\mathrm{report}}
+
\lambda_{\mathrm{shape}}\mathcal{L}_{\mathrm{shape}}.
\end{aligned}
\end{equation}
\(\mathcal{L}_{\mathrm{LM}}\) is masked
cross-entropy over the structured assistant output.
\(\mathcal{L}_{\mathrm{lesion}}\) and \(\mathcal{L}_{\mathrm{write}}\)
are binary cross-entropies on per-step lesion presence and
memory-write targets, with positive-class weighting on the write loss.
\(\mathcal{L}_{\mathrm{report}}\) supervises episode-level polyp/ulcer
and stenosis presence at the final chunk.
\(\mathcal{L}_{\mathrm{shape}}\) is applied only when geometry is available. When without geometry, the language-model target for \(o_k\) is \textit{not observable}. Samples labeled \textit{needs review} are down-weighted by \(\lambda_{\mathrm{rev}}\)
in the language, lesion, and write losses.

Stage~1 uses stride-four samples and the objective above.
Stage~2 continues from the Stage~1 checkpoint to refine
reconstruction control. It keeps reconstruction-transition frames in
addition to stride-four samples and feeds the preceding target
controller state as input. Motion-action tokens receive weight
\(\omega_{\mathrm{mot}}\), \textit{activation/deactivation} tokens
\(\omega_{\mathrm{act}}\), and reconstruction \textit{no change}
tokens \(\omega_{\mathrm{nc}}\).

\subsection{Inference}
\label{sec:inference}
At test time each decision uses only current and past images, the
task instruction, the executed reconstruction state, and any
completed geometry. Ground-truth reasoning, navigation labels, and
reconstruction events are not provided. The language head generates
\(\mathbf{Y}_k\), where a second forward pass over the prompt
updates memory and the report heads. The episode ends when the VLA
predicts a return to the normal lumen and emits \textit{stop}.

The executed reconstruction state is set from the parsed navigation
state \(n_k\), not from \(\hat{a}^r_k\). Reconstruction is switched
ON for \textit{prepare stenosis entry} and \textit{traverse stenosis},
and OFF for \textit{confirm stenosis end}, \textit{completed}, and
\textit{terminated}. Otherwise, the previous state is kept.
DA3 runs asynchronously with the VLA.
Only completed batches from the active interval are consumed, and
pending work is cancelled at episode end.
The report is assembled as in Sec.~\ref{sec:outputs}.

\section{Experiments}

\subsection{Datasets}
\label{sec:exp-data}
We use the EndoCausal dataset as described in Sec.~\ref{sec:data}. The dataset includes 400 navigation episodes, about 0.8 million frames at 20\,fps, collected in anatomically structured esophageal and colonic phantoms. Episodes are split into 320 training, 40 validation, and 40 test sequences. The augmented episodes cover lumen exploration, lesion inspection, stenosis approach, alignment, entry, traversal, and termination. Training-only augmentation produces approximately 2{,}200 additional
sequences via brightness, contrast, colour, blur, horizontal and
vertical flips, and rotations of up to \(45^\circ\). Images, masks,
convex hulls, spatial descriptions, and directional actions are
transformed consistently. Their point maps are excluded and
geometry is marked unavailable. Validation, test, and physical-robot
trials are unaugmented.

\subsection{Implementation Details}
Stage~1 runs for three epochs and Stage~2 for one epoch.
The loss weights are
\(\lambda_{\mathrm{LM}}=\lambda_{\mathrm{report}}=1\),
\(\lambda_{\mathrm{lesion}}=0.5\),
\(\lambda_{\mathrm{write}}=0.25\),
and \(\lambda_{\mathrm{shape}}=0.1\).
Review samples use \(\lambda_{\mathrm{rev}}=0.5\).
In Stage~2, token weights are
\(\omega_{\mathrm{mot}}=3\),
\(\omega_{\mathrm{act}}=20\),
and \(\omega_{\mathrm{nc}}=1\).
Training uses AdamW with learning rates of \(5\times10^{-6}\) for LoRA
and \(5\times10^{-5}\) for other modules. We apply the weight decay \(0.01\), and gradient clipping at norm 1.0 for all trainable parameters. Memory is
propagated across chunks of the same episode, detached at chunk
boundaries, and reset between episodes. Decoding is greedy with at most 192 new tokens and KV caching disabled. When reconstruction is ON, up to 16 views are sampled with stride two. A new DA3 request is submitted only if the worker is idle and at least five decisions have elapsed since the previous submission in the active interval. All experiments are conducted using an NVIDIA
RTX A6000 GPU.

\subsection{Evaluation protocol}
\label{sec:metrics}
Table~\ref{tab:main} reports four quantities under two protocols.
Semantic and Action are open-loop scores on the 40 held-out recorded
EndoCausal test episodes, using Sec.~\ref{sec:inference} at stride four.
Esophagoscopy (Esophag.) and colonoscopy (Colon.) are closed-loop success rates of the same policies on the physical 3-DoF endoscope in esophageal and colonic phantoms. Avg.\ is the unweighted mean of these four metrics.
Ablation and zero-shot tables use the same recorded test set and the same physical-trial counts.\\

\noindent\textbf{Semantic.}
Scored open-loop on the 40 recorded test episodes (not on the physical robot).
Each generated string is parsed into six tags:
\{\textit{reasoning}, \textit{navigation\_state},
\textit{safety\_constraint}, \textit{opening\_shape},
\textit{motion\_action}, \textit{reconstruction\_action} \}.
The Semantic score covers the parsed \textit{reasoning} field, the
per-step \(o_k\) in \(\mathbf{Y}_k\), and the episode-level lesion report
(opening shape \(o_{k^\star}\) in Sec.~\ref{sec:outputs}).
It is the fraction of strings that exactly match the annotated targets after removing whitespace. Methods without this are marked `--'.\\

\noindent\textbf{Action.}
On the same 40 recorded episodes and decisions, the fraction whose parsed \textit{motion\_action}
exactly equals the annotated discrete command \(a_k\in\mathcal{A}\).\\

\noindent\textbf{Task success.}
We also evaluate the real-world deployment on the physical 3-DoF endoscope in the same esophageal and colonic
phantoms used for collection, from a fixed entrance pose.
A trial counts as success only if every condition holds 1) any available
polyp or ulcer is detected; 2) the stenosis is identified and kept in the
field of view; 3) polyp contact is avoided; the policy emits \textit{stop}
after returning to the normal lumen; 4) and the episode-level report is correct.
Report correctness uses the memory heads at threshold \(0.5\) for polyp, ulcer, and stenosis presence, and opening shape \(o_{k^\star}\).
Failures are labelled by the primary cause: incorrect lesion or stenosis
detection, incorrect reasoning, navigation failure, or incorrect
reporting.
Each policy is evaluated on 36 independent physical trials per procedure (36 esophagoscopy and 36 colonoscopy).

\begin{figure*}[t]
    \centering
    \includegraphics[width=1.0\textwidth]{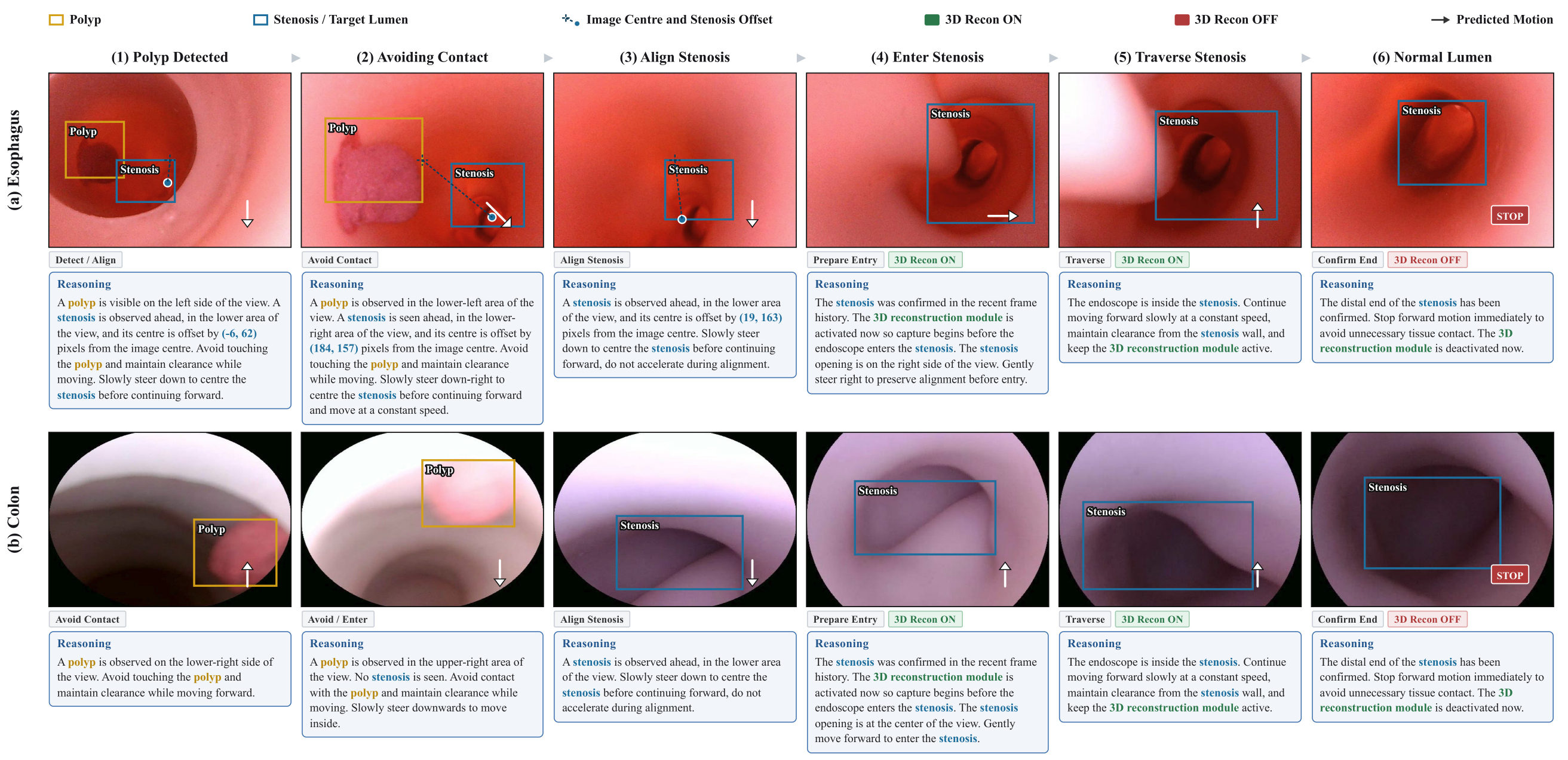}
    \caption{Qualitative reasoning traces of StenoVLA-3D on esophagus (a) and colon (b) sequences. Columns (1)–(6) follow polyp detection, contact avoidance, stenosis alignment, reconstruction-gated entry, traversal, and distal-end stop. Yellow boxes mark polyps. Cyan boxes mark the stenosis or target lumen. Green/red badges indicate whether 3D reconstruction is active.}
    \label{fig_main}
\end{figure*}

\subsection{Baselines}
We compare StenoVLA-3D against two baselines. EndoVLA$^*$ is adapted from EndoVLA~\cite{ng2025endovla}, a language-conditioned endoscopic tracking VLA with a 2-DoF action space, by expanding the actions to 3-DoF to match our robotic endoscope. For a controlled comparison, EndoVLA$^{*}$ receives the same current and historical frames, image resolution, task instruction, dataset split, and 3-DoF action space as StenoVLA-3D. $\pi_{0.5}$~\cite{intelligence2025pi05visionlanguageactionmodelopenworld} is a generalist VLA: it predicts continuous actions with a flow-matching action expert. For comparison with our discrete action labels, each continuous motor-command vector predicted by $\pi_{0.5}$ is mapped to the closest canonical motor-command vector used by our controller. Commands below a validation-selected deadband $\tau$ are mapped to \texttt{stop}. Otherwise, cosine similarity determines the corresponding forward, backward, steering, or diagonal action. The resulting labels are evaluated using the same exact-match
action metric as StenoVLA-3D. Both of them lack endoscopic geometry fusion and memory components.
We also evaluate two 3D-aware variants, Steno-DAC and Steno-VGGT, obtained by replacing DA3 with EndoDAC and VGGT, respectively, while keeping the same control logic, memory mechanism, and evaluation protocol. This design enables a fair assessment of both the complete StenoVLA-3D framework and the contribution of its DA3-based geometric representation.

\begin{table}[!t]
\centering
\caption{Performance comparison of baselines and StenoVLA-3D variants. Semantic and Action are open-loop scores on 40 recorded test episodes. Esophag.\ and Colon.\ are physical-robot success (\%) over 36 trials per procedure. EndoVLA$^*$ is adapted from EndoVLA~\cite{ng2025endovla} by using Qwen3-VL~\cite{bai2025qwen3} and replacing the action space from 2-DoF to 3-DoF.}
\label{tab:main}
\resizebox{\linewidth}{!}{
\begin{tabular}{lccccc}
\toprule
\textbf{Method} &  \textbf{Semantic} & \textbf{Action} & \textbf{Esophag.} & \textbf{Colon.}  & \textbf{Avg.} \\
\midrule
EndoVLA$^*$~\cite{ng2025endovla}  & 80.2 & 67.3 & 50.0 & 55.6 & 63.3 \\
$\pi_{0.5}$~\cite{intelligence2025pi05visionlanguageactionmodelopenworld}  & -- & 55.2 & 33.3 & 22.2 & 36.9 \\
\midrule
\multicolumn{6}{l}{\textit{StenoVLA-3D variants}} \\
\midrule
Steno-DAC     & 92.1 & \textbf{84.7} & 63.9 & 52.8 & 73.4 \\
Steno-VGGT  & 88.8 & 81.0 & 58.3 & 66.7 & 73.7 \\
\textbf{StenoVLA-3D}  & \textbf{95.2} & {83.4} & \textbf{88.9} & \textbf{77.8} & \textbf{86.3} \\
\bottomrule
\end{tabular}
}
\end{table}

\begin{table}[!t]
\centering
\caption{Architectural components and control frequency. Discrete methods emit one motion action every four frames on 20\,fps recordings. $\pi_{0.5}$ executes 50-step action chunks at 50\,Hz. EndoVLA$^*$ is modified from EndoVLA~\cite{ng2025endovla} with the action expanded from 2-DoF to 3-DoF.}
\label{tab:method_comparison}
\resizebox{\linewidth}{!}{
\begin{tabular}{@{}lcccc@{}}
\toprule
\textbf{Method} & \textbf{Persistent} & \textbf{Action} & \textbf{Geometry}  & \textbf{Hz}\\
 & \textbf{memory} & \textbf{space} & \textbf{backbone}  &  \\
\midrule
EndoVLA$^*$~\cite{ng2025endovla}
    & $\times$ & discrete & -- & $5$ \\
$\pi_{0.5}$~\cite{intelligence2025pi05visionlanguageactionmodelopenworld}
    & $\times$ & continuous & --  & $50$ \\
\midrule
\multicolumn{5}{@{}l}{\textit{StenoVLA-3D variants}} \\
\midrule
Steno-DAC
    & $\checkmark$ & discrete & EndoDAC & $5$ \\
Steno-VGGT
    & $\checkmark$ & discrete & VGGT  & $5$ \\
\textbf{StenoVLA-3D}
    & $\checkmark$ & discrete & DA3  & $5$ \\
\bottomrule
\end{tabular}
}
\end{table}

\subsection{Quantitative Comparison with Baselines and Variants}
Table~\ref{tab:main} compares StenoVLA-3D with EndoVLA$^*$ and $\pi_{0.5}$. StenoVLA-3D achieves the highest performance across all reported metrics, obtaining 95.2 in semantic score, 83.4 in action prediction, 88.9\% in esophagoscopy, and 77.8\% in colonoscopy, with an overall average of 86.3. Compared with EndoVLA$^*$, this corresponds to improvements of 15.0, 16.1, 38.9, and 22.2 percentage points, respectively. StenoVLA-3D also substantially outperforms $\pi_{0.5}$ in action prediction and both physical procedures.

\begin{figure*}[t]
    \centering
    \includegraphics[width=1.0\textwidth]{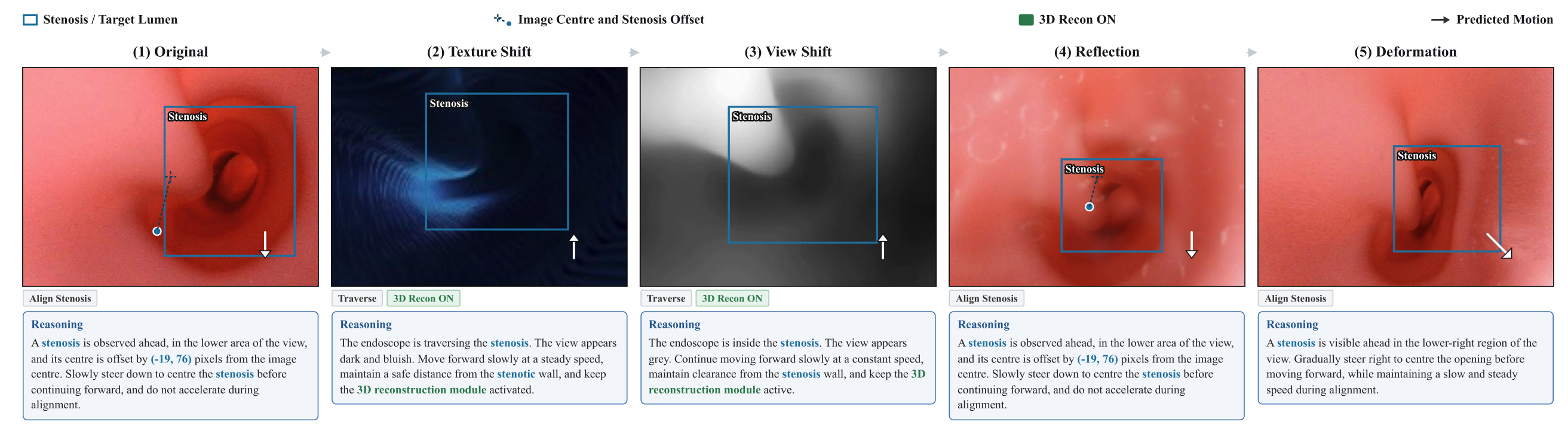}
    \caption{Zero-shot qualitative results of StenoVLA-3D. Column (1) is the unperturbed view; (2)–(5) show texture shift, view shift, reflection, and deformation. Cyan boxes mark the stenosis or target lumen. The model retains consistent reasoning and navigation under these appearance and geometry changes.}
    \label{zero_shot}
\end{figure*}
Among the 3D-aware variants, Steno-DAC and Steno-VGGT achieve average scores of 73.4\% and 73.7\%, compared with 86.3\% for StenoVLA-3D. Although Steno-DAC obtains a slightly higher action score than StenoVLA-3D (84.7\% versus 83.4\%), StenoVLA-3D performs considerably better in semantic score and physical endoscopic navigation. The physical gap is largest in esophagoscopy (88.9\% versus 63.9\% and 58.3\%), indicating that DA3 geometry transfers more reliably to stenosis traversal than EndoDAC or VGGT.

Table~\ref{tab:method_comparison} summarizes architectural components and control frequency.
On 20\,fps recordings the discrete policies act every fourth frame, i.e., one motion action per four frames (\(20/4=5\)\,Hz in video time), matching the stride used at training and inference.
\(\pi_{0.5}\) instead executes a 50-step action chunk at 50\,Hz.
This 5\,Hz rate is the command schedule, not wall-clock throughput: DA3 geometry estimation required approximately \(3.63\,\mathrm{s}\) per reconstruction job, while the VLA required \(25.20\,\mathrm{s}\) to generate a decision, for \(28.83\pm0.65\,\mathrm{s}\) per decision overall.
Steno-DAC and Steno-VGGT use the same discrete-action schedule and persistent-memory design but replace DA3 with EndoDAC or VGGT, with wall-clock latencies of \(32.71\pm0.25\,\mathrm{s}\) and \(36.85\pm0.78\,\mathrm{s}\) per decision.
EndoVLA$^*$ has neither persistent memory nor a geometry backbone, yet is slower (\(35.12\pm0.57\,\mathrm{s}\)) than StenoVLA-3D.
\(\pi_{0.5}\) is the fastest at \(1.45\pm0.41\,\mathrm{s}\) per inference because it predicts continuous actions without language reasoning or geometry. This speed does not translate into comparable action or procedural success in Table~\ref{tab:main}.

\begin{table}[t]
\centering
\caption{Ablation of StenoVLA-3D components under the same protocol as Table~\ref{tab:main}: Semantic/Action on 40 recorded test episodes. Esophag./Colon.\ physical-robot success over 36 trials per procedure (\%).}
\label{tab:ablation}
\resizebox{\linewidth}{!}{
\begin{tabular}{lccccc}
\toprule
\textbf{Method} &  \textbf{Semantic} & \textbf{Action} & \textbf{Esophag.} & \textbf{Colon.}  & \textbf{Avg.} \\
\midrule
w/o reasoning     & -- & 56.5 & 47.2 & 41.7 & 48.5 \\
w/o persistent memory  & 88.0 & 79.2 & 0.0 & 0.0 & 41.8 \\
w/o temporal memory & 72.3 & 63.1 & 38.9 & 30.6 & 51.2 \\
w/o geometry fusion  & 85.5 & 78.3 & 0.0 & 0.0 & 41.0 \\
\textbf{StenoVLA-3D}  & \textbf{95.2} & \textbf{83.4} & \textbf{88.9} & \textbf{77.8} & \textbf{86.3} \\
\bottomrule
\end{tabular}
}
\end{table}
Fig.~\ref{fig_main} shows step-level reasoning on esophagus (a) and colon (b) sequences. Columns (1)--(6) follow polyp detection, contact avoidance, stenosis alignment, reconstruction-gated entry, traversal, and distal-end stop. In both organs the policy names the polyp when it is visible, measures the stenosis offset from the image centre, and selects an 8-connected steering action that reduces that offset while keeping clearance. Reconstruction is switched on at confirmed entry (green) and off after the distal lumen is recognized (red), matching the navigation-state controller in Sec.~\ref{sec:inference}. The colon sequence additionally shows a stage in which a polyp is observed before the stenosis enters the field of view, after which alignment and gated reconstruction proceed as in the esophagus.

\begin{table}[t]
\centering
\caption{Zero-shot performance under unseen visual domain shifts, using the same protocol as Table~\ref{tab:main} (40 recorded test episodes, 36 physical trials per procedure). Semantic and Action are recorded open-loop scores. Esophag. and
Colon. denote the episode-level success on prerecorded esophageal and colonic sequences, respectively.}
\label{tab:zero-shot}
\resizebox{\linewidth}{!}{
\begin{tabular}{lccccc}
\toprule
\textbf{Method} &  \textbf{Semantic} & \textbf{Action} & \textbf{Esophag.} & \textbf{Colon.}  & \textbf{Avg.} \\
\midrule
Texture Shift & 77.0 & 62.2 & 55.6 & 50.0 & 61.2 \\
View Shift  & 91.1 & 77.8 & 83.3 & 72.2 & 81.1 \\
Deformation & 66.1 & 63.9 & 69.4 & 63.9 & 65.8 \\
Reflection  & 90.7 & 81.0 & 77.8 & 66.7 & 79.1 \\
\bottomrule
\end{tabular}
}
\end{table}

\subsection{Ablation study}
Table~\ref{tab:ablation} isolates four components. Removing reasoning supervision reduces the overall average from 86.3\% to 48.5\%, with action decreasing from 83.4\% to 56.5\% and success in esophagoscopy and colonoscopy declining from 88.9\% and 77.8\% to 47.2\% and 41.7\%, respectively. Without temporal memory, the semantic score falls to 72.3\%, action to 63.1\%, and the average to 51.2\%, indicating that sequential context is needed for navigation-state estimation and timely DA3 activation. Removing persistent memory leaves step-level scores relatively high (88.0\% semantic, 79.2\% action) but yields 0\% procedural success, because the episode-level report cannot retain lesions after they leave the view. Removing geometry fusion similarly preserves much of the open-loop scores (85.5\% semantic, 78.3\% action) while collapsing both procedures to 0\%, consistent with the need for point-map features in stenosis-shape estimation and 3D-aware closed-loop navigation. Overall, reasoning and temporal memory affect both recorded-episode and physical scores, whereas persistent memory and geometry fusion are required for task success under the reporting protocol.

\subsection{Cross-Domain Evaluation}
Table~\ref{tab:zero-shot} evaluates zero-shot generalization under four unseen visual domain shifts, without additional fine-tuning. View shift is closest to the in-domain result in Table~\ref{tab:main}, with 91.1\% semantic, 77.8\% action, 83.3\% esophagoscopy, and 72.2\% colonoscopy. Reflection gives the second-highest average score, with 90.7\% semantic accuracy and 81.0\% action accuracy, although colonoscopy success decreases to 66.7\%.
 Deformation reduces reasoning the most (66.1\%) and lowers the average to 65.8\%, while texture shift is the weakest overall, with colonoscopy success falling to 50.0\%. Viewpoint changes are comparatively well tolerated, whereas surface-texture changes and geometric deformation remain limited. Fig.~\ref{zero_shot} shows a representative stenosis view under the same four perturbations. Across columns, the policy still localizes the lumen, reports a consistent opening-centred interpretation, and issues a 3D-aware navigation decision.

\subsection{Limitations}
The current implementation is limited by inference latency. VLA generation requires approximately \(25.2\,\mathrm{s}\) per decision,
while each asynchronous DA3 reconstruction job requires \(3.63 \pm 0.02\,\mathrm{s}\). Although reconstruction does not block
VLA execution, generation remains the main computational bottleneck and prevents continuous low-level control. Further limitations include phantom-only validation, the absence of calibrated stenosis-length and
lumen-diameter estimates, and the mismatch between jointly reconstructed offline training geometry and causal online geometry. 

\section{Conclusion}
We presented StenoVLA-3D, a 3D-aware VLA framework for robotic stenosis navigation and episode-level reporting, together with the
EndoCausal dataset. By combining point-map-conditioned prediction, navigation-controlled reconstruction, and memory-based reporting,
StenoVLA-3D achieved task success rates of \(88.9\%\) and \(77.8\%\) in esophageal and colonic phantoms, respectively. Ablation and
cross-domain evaluations further demonstrated the contributions of 3D information and memory to reasoning, navigation, and reporting. Future work will investigate quantization, KV caching, shorter outputs, hardware-aware optimization, causal-window retraining, and calibrated metric geometry.

\bibliographystyle{ieeetr}
\bibliography{references}
\end{document}